\documentclass[journal,twoside,web]{ieeecolor}
\usepackage{generic}
\usepackage{cite}
\usepackage{amsmath,amssymb,amsfonts}
\usepackage{algorithmic}
\usepackage{graphicx}
\usepackage{textcomp}
\usepackage[x11names]{xcolor}
\usepackage{multirow}
\usepackage[english,main=english]{babel}
\usepackage{lipsum}
\usepackage{caption}
\usepackage{float}

\def\BibTeX{{\rm B\kern-.05em{\sc i\kern-.025em b}\kern-.08em
    T\kern-.1667em\lower.7ex\hbox{E}\kern-.125emX}}
\begin{document}

\title{Img2CAD: Conditioned 3D CAD Model Generation from Single Image with Structured Visual Geometry}



\author{Tianrun Chen, Chunan Yu, Yuanqi Hu, Jing Li, Tao Xu, Runlong Cao, Lanyun Zhu, Ying Zang, Yong Zhang, Zejian Li, Linyun Sun
\thanks{T. Chen and C. Yu contributed equally to this research}
\thanks{T. Chen is with the College of Computer Science and Technology, Zhejiang University and KOKONI3D, Moxin (Huzhou) Technology Co., LTD, China. E-mail: tianrun.chen@zju.edu.cn}
\thanks{Z. Li is with the School of Software Technology, Zhejiang University, China}
\thanks{L. Sun is with the College of Computer Science and Technology, Zhejiang University, China}
\thanks{C. Yu, Y. Hu, J. Li, T. Xu, R. Cao, Y. Zang, and Y. Zhang are with the School of Information Engineering, Huzhou University, China}
\thanks{L. Zhu is with the Information Systems Technology and Design Phillar, Singapore University of Technology and Design (SUTD), Singapore.}}

\maketitle
\vspace{-3cm} 

\renewcommand\twocolumn[1][]{#1}
\begin{figure*}[!ht]
    \centering
    \captionsetup{type=figure}
    \includegraphics[width=0.8\textwidth]{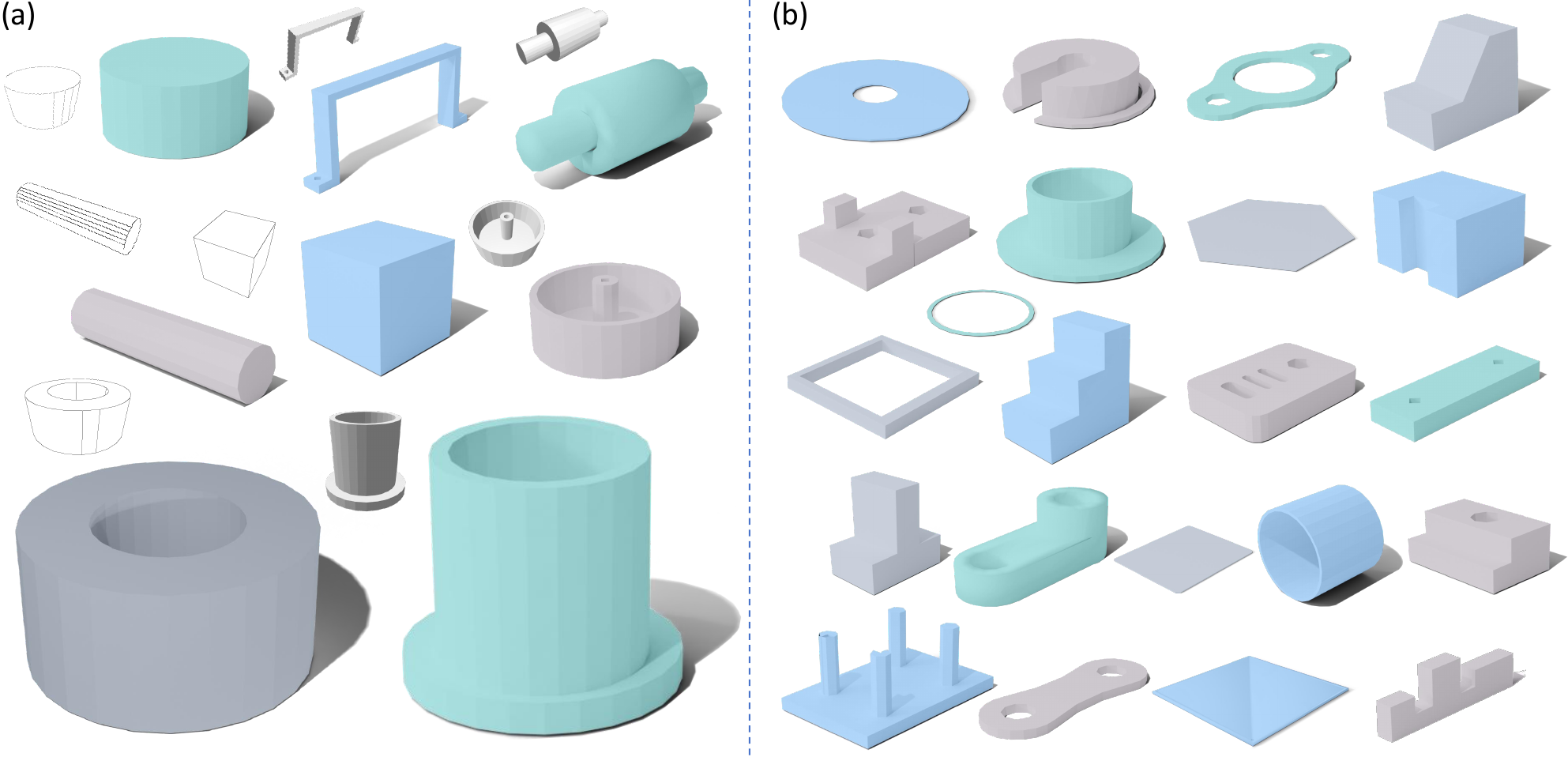}
    \caption{We propose the first 3D model generation network that can produce “sketch and extrude" parametric command representation of 3D objects with only single images or sketches as the input. (a) Examples generated from the images or sketches. (b) More examples of CAD designs were obtained from our Img2CAD approach, which can be used as the coarse rapid prototyping stage for 3D modeling by experts to make the modeling process faster.}
\end{figure*}

\begin{abstract}
   In this paper, we propose Img2CAD, the first approach to our knowledge that uses 2D image inputs to generate CAD models with editable parameters. Unlike existing AI methods for 3D model generation using text or image inputs often rely on mesh-based representations, which are incompatible with CAD tools and lack editability and fine control, Img2CAD enables seamless integration between AI-based 3D reconstruction and CAD software. We have identified an innovative intermediate representation called Structured Visual Geometry (SVG), characterized by vectorized wireframes extracted from objects. This representation significantly enhances the performance of generating conditioned CAD models. Additionally, we introduce two new datasets to further support research in this area: ABC-mono, the largest known dataset comprising over 200,000 3D CAD models with rendered images, and KOCAD, the first dataset featuring real-world captured objects alongside their ground truth CAD models, supporting further research in conditioned CAD model generation. 
\end{abstract}


\section{Introduction}
    3D modeling has numerous applications across industries, ranging from product design and architecture to animation and immersive virtual environments. Traditionally, creating high-quality 3D models has been the domain of skilled professionals, requiring extensive knowledge of complex software and a steep learning curve. However, with the advent of the AI generative model, there is growing optimism that these barriers can be lowered, making 3D content creation more accessible to a wider audience. 
    
    However, despite the remarkable advancements in generative models \cite{SyncDreamer, Wonder3D, AutoSDF, Slice3D, tang2023make, shi2023mvdream, zhang2019view} that can generate diverse 3D models from intuitive input like images or text, we identified a significant gap between current 3D modeling approaches and their practical applications, particularly in the field of fabrication. Specifically, most man-made objects are initially designed using Computer-Aided Design (CAD) software in a parametric form, yet existing 3D AI-generated content (3D AIGC) algorithms predominantly rely on mesh-based representations. 

    A key limitation of mesh-based representations used by current AI 3D generation approaches lies in interpretability and editability. Unlike CAD's parametric design, which allows users to easily modify specific features, mesh-based models are often difficult to edit with precision, limiting user control and flexibility. Furthermore, the surface quality and compactness of these generated meshes frequently fall short, especially when produced using algorithms like Marching Cubes, which convert signed distance functions (SDF) into mesh form. This process often results in surfaces that are not fully smooth and edges that are insufficiently sharp, making the models less suitable for applications such as relightable rendering and animation \cite{chen2020bspnet, UCSG_Net}. These geometric imperfections can be further exacerbated when the models are combined with certain materials. In contrast, CAD's parametric models offer higher precision and flexibility. Parametric design allows users to directly modify a model's geometry through parameters, offering greater interpretability and enabling rapid, precise adjustments. This brings us to an important question: \textit{If CAD models offer such clear advantages, why do most current 3D AIGC methods focus on mesh-based generation?}
    
    We believe that there are two main reasons for this. First, most large-scale 3D datasets \cite{chang2015shapenet,deitke2023objaverse,deitke2024objaverse} available today are mesh-based, which provides a wealth of diverse models for training. These mesh-based datasets have been instrumental in driving the progress of 3D AIGC, but they limit the ability to directly generate CAD-compatible content. Second, there is a significant domain gap between input formats like images or text and the structure of CAD models. A CAD model is composed of a sequence of geometric operations, such as curve sketching, extrusion, filleting, boolean operations, and chamfering, each governed by specific parameters \cite{hoffmann2001towards}. Some of these parameters are discrete options, while others are continuous values. To generate a valid CAD model, a network must accurately learn both the sequence of operations and the associated values, a task complicated by the fact that incorrect formats can result in invalid outputs that cannot be parsed by CAD kernels, leading to complete generation failures. 
    
    The challenge is further amplified when working with in-the-wild images, which may feature varied lighting, backgrounds, and perspectives. Reconstructing a CAD model from such images is extremely difficult because a single image often only captures part of an object, requiring the model to generate the unseen portions. This process demands a wealth of prior knowledge, but no existing CAD dataset is as diverse or extensive as mesh-based datasets. Thus, while CAD models hold immense potential for improving the quality and applicability of 3D AIGC, overcoming these obstacles will require further advancements in both datasets and models. 

    Given the significant challenges in generating CAD models, existing approaches have primarily focused on training intelligent agents to reconstruct CAD models from point clouds. Few methods have ventured into CAD model generation itself, and those that do often focus on unconditioned generation or are limited to coarse conditioning, such as category information. This leaves considerable room for innovation in controlled, conditional CAD model generation using more detailed inputs like images. By addressing the challenges of generating CAD models directly from these rich inputs, we could unlock new possibilities in precise and flexible 3D content creation, paving the way for more accessible and practical applications in industries like fabrication and design.

    This work aims to address the existing research gap in CAD model generation. To the best of our knowledge, we propose the first single image-conditioned CAD generation network, Img2CAD, which outputs a sequence of sketch and extrusion operations. These operations can be parsed by a CAD kernel to produce a Boundary Representation (B-Rep) format, enabling seamless integration into existing CAD software. To further support research in this area, we introduced two new datasets: \textbf{ABC-mono}, the largest dataset comprising over 200,000 3D CAD models paired with rendered images, and \textbf{KOCAD}, the first dataset featuring real-world captured objects fabricated by addictive manufacturing alongside their corresponding ground truth CAD models. These datasets are designed to promote advances in controlled conditional CAD generation from diverse inputs, bridging the gap between AI-driven modeling and practical industrial applications.

\begin{figure}[t]
\centering
\includegraphics[width=0.5\textwidth]{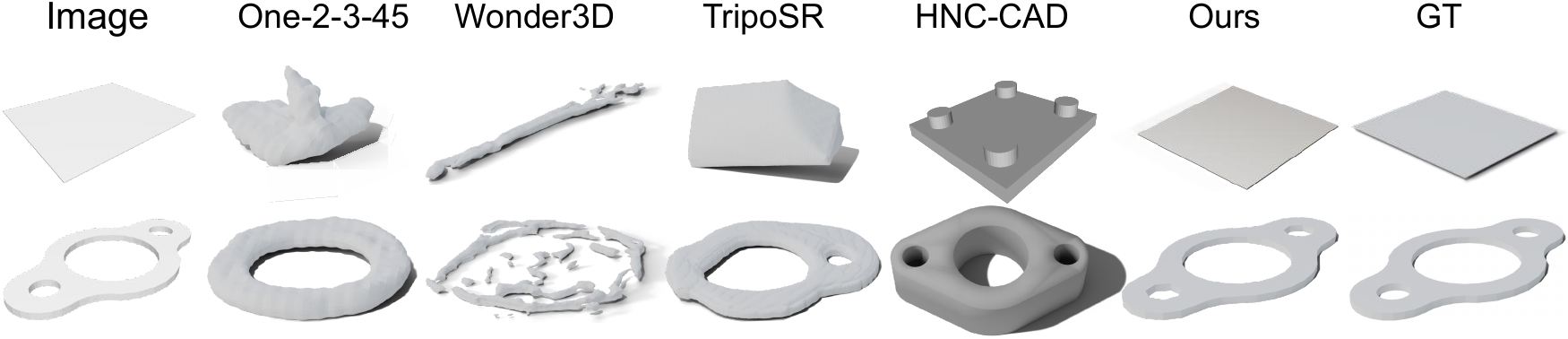}
\caption{The Limitation of Existing 3D AIGC Approach in Generating Simple Man-Made Geometry with Image Conditions. Our method aims to generate high-quality man-made objects guided by image conditions using CAD representation instead of using mesh, which allows high-quality surface creation and direct integration with traditional workflow.} \label{Fig000}
\vspace{-0.5cm}
\end{figure}

\begin{table*}[!htbp]
\caption{Comparison between this work and other 3D AIGC approaches.}
\resizebox{\textwidth}{!}{
\begin{tabular}{|c|c|c|c|c|c|c|c|c|}
\hline
\begin{tabular}[c]{@{}c@{}} Representative \\ Works \end{tabular} & \begin{tabular}[c]{@{}c@{}} \cite{NIPS2019_8340, Slice3D, TripoSR2024, Jun_Nichol, liu2023one2345, chen2020bspnet, wang2024crm} \end{tabular}
& \begin{tabular}[c]{@{}c@{}} \cite{yu2021pixelnerf, gu2023nerfdiff, mueller2022autorf, melaskyriazi2023realfusion} \end{tabular}
& \begin{tabular}[c]{@{}c@{}} \cite{kong2022diffusion,  Luo_Hu_2021, Cheng_Chai_Ren_Lee_Olszewski_Huang_Maji_Tulyakov_2022} \end{tabular} 
& \begin{tabular}[c]{@{}c@{}} \cite{Li_Wang_Tseng_2023, EnVision2023luciddreamer, tang2024lgm, tang2023dreamgaussian, xu2024grm} \end{tabular} 
& \begin{tabular}[c]{@{}c@{}} \cite{Jayaraman_Lambourne_Desai_Willis_Sanghi_Morris,  Xu2024brepgen, GuoComplexGen2022} \end{tabular}
& \begin{tabular}[c]{@{}c@{}} \cite{ren2021csg, Yu_Chen_Tanveer_Amiri_Zhang_2023, CAPRI-Net, UCSG_Net,  Sharma_Goyal_Liu_Kalogerakis_Maji_2018} \end{tabular} 
& \begin{tabular}[c]{@{}c@{}} \cite{Li_Guo_Zhang_Yan_2023, Xu_Peng_Cheng_Willis_Ritchie_2021,  ExtrudeNet,Wu_Chang_Zheng_2021, HNC_CAD,Xu_Willis_Lambourne_Cheng_Jayaraman_Furukawa_2022,  Zhou_Tang_Zhou_2023} \end{tabular} & {\color[HTML]{009901} \textbf{Ours}} \\ \hline
Category & \begin{tabular}[c]{@{}c@{}} 3D Mesh/Voxel \end{tabular} & \begin{tabular}[c]{@{}c@{}} Neural Fields \end{tabular} & \begin{tabular}[c]{@{}c@{}} Point Cloud \end{tabular} & \begin{tabular}[c]{@{}c@{}} Gaussian Splatting \end{tabular} & \begin{tabular}[c]{@{}c@{}}  Direct B-Rep \end{tabular} & \begin{tabular}[c]{@{}c@{}} Constructive Solid Geometry \end{tabular} & \begin{tabular}[c]{@{}c@{}c@{}} Construction Sequence Generation \end{tabular} & {\color[HTML]{009901} \textbf{Ours}} \\ \hline
Input Condition & \begin{tabular}[c]{@{}c@{}}Image/Text \end{tabular} & \begin{tabular}[c]{@{}c@{}} Image/Text \end{tabular} & \begin{tabular}[c]{@{}c@{}} Image/ \\Text \end{tabular} & \begin{tabular}[c]{@{}c@{}} Image/Text \end{tabular} & \begin{tabular}[c]{@{}c@{}} Unconditioned/Class Label \end{tabular} & \begin{tabular}[c]{@{}c@{}c@{}} Unconditioned/Class/Voxel/Point Cloud\end{tabular} & Unconditioned & {\color[HTML]{009901} \begin{tabular}[c]{@{}c@{}} \textbf{Single Image/} \\ \textbf {Sketch} \end{tabular}} \\ \hline
Editability & Low & Low & Low & Low & \textbf{High} & Medium & \textbf{High} & {\color[HTML]{009901} \textbf{High}} \\ \hline
Interpretability & Low & Low & Low & Low & Medium & Medium & \textbf{High} & {\color[HTML]{009901} \textbf{High}} \\ \hline
Compactness & Low & Low & Low & Low & \textbf{High} & \textbf{High} & \textbf{High} & {\color[HTML]{009901} \textbf{High}} \\ \hline
\end{tabular}}
\vspace{-0.4cm}
\label{compt}
\end{table*}

    Specifically, we adopt a Transformer-based network to encode the image input and another Transformer to decode the command and parameter outputs. To address the substantial domain gap between images and CAD models, we introduce a novel intermediate representation for CAD reconstruction: Structured Visual Geometry (SVG). This representation is a vectorized wireframe \cite{xue2023holistically,xue2020holistically,ma20223d} of the object, derived from a geometric parser. The SVG explicitly extracts line segment and joint representations, which serve as crucial guides for reconstructing the CAD sequence. We employ the Holistic Attention Transformer (HAT) field to encode the line segments through a closed-form 4D geometric vector field, which generates dense sets of line segments and extracts endpoint proposals from heatmaps. These dense line segments are bound with sparse endpoint proposals to form initial wireframes. To further refine the results, we introduce the Joint-Decoupled Line-of-Interest Aligning (JD LOIAlign) module, which filters out false positive proposals through interest point alignment. This module captures the co-occurrence between the endpoint proposals and the HAT field, enhancing data optimization. After these operations, we fuse both types of data using cross-attention and feed them into the decoder, which produces the command types and parameters for CAD generation.

    We conducted extensive experiments, achieving state-of-the-art (SOTA) performance in terms of fidelity, surface quality, and inference speed compared to existing popular 3D AIGC methods that use images as input. Our approach is robust to input variations due to the explicit structured visual geometry understanding it incorporates and can generate high-fidelity and high-quality 3D CAD models from both sketch input and in-the-wild images. We also showcased a downstream application of our method: adding materials to the generated CAD models for rendering. Moreover, we believe that the CAD models generated by our method can serve as an excellent starting point for professional designers. Designers are often trained to model from "rough to fine," and our system efficiently completes the initial "rough" prototyping stage, allowing designers to focus on refining and adding intricate details in subsequent steps.

    While the current complexity of datasets poses certain limitations, we plan to explore more complex data and advanced modeling steps in future work. Nevertheless, we are confident that this work marks a significant first step towards bridging the gap between AI-generated CAD models and real-world applications, offering a promising foundation for further research and innovation in the field. In summary, our contributions are the following:
    
    \begin{itemize}
        \item We expand the representation of the existing 3D AIGC method and propose the first end-to-end image-conditioned 3D CAD model generation method that produces \textbf{sketch and extrude sequences compatible with existing CAD software} and has good \textbf{editability, interpretability, and compactness} (For full comparison, refer to Table. \ref{compt}).
        \item Our research demonstrates the effectiveness of structured visual geometry understanding as a powerful tool for enhancing the performance of image-conditioned 3D CAD model generation.
        \item We create a new dataset, ABC-mono, for image-conditioned CAD command generation. This new dataset is an extended version of the existing ABC dataset, enriched with a total of more than 200,000 valid human-designed 3D models and more than 15,000,000 of their corresponding image and sketch pair, making it the largest dataset of its kind known to date.
        \item We create a new in-the-wild dataset for image and CAD model pair, KOCAD. By fabricating the models and capturing images of the objects in varying conditions, we set a new evaluation benchmark for challenging in-the-wild CAD model generation with the KOCAD dataset.
    \end{itemize}

\section{Related Works}
\noindent{\textbf{3D Generation and Image Conditioning. }}
\label{ssec:2.1}
    Single-view image reconstruction, driven by the rapid advancements in generative models \cite{SyncDreamer, Wonder3D, AutoSDF, Slice3D}, has emerged as a vibrant research domain. Table. \ref{compt} shows the full comparison between our work and some representative 3D generation works.

\noindent{\textbf{Parametric Inference. } }
\label{ssec:2.2}
    Parametric shape inference has seen significant advancements recently, enabling neural networks to analyze geometric data and infer parametric shapes. For instance, ParSeNet \cite{Sharma_Liu_Maji_Kalogerakis_Chaudhuri_Mech_2020} decomposes a 3D point cloud into a set of parametric surface patches, UV-Net \cite{Jayaraman_Sanghi_Lambourne_Davies_Shayani_Morris_2020} and BrepNet \cite{Lambourne_Willis_Jayaraman_Sanghi_Meltzer_Shayani_2021} concentrate on encoding the boundary curves and surfaces of parametric models. Xu et al. \cite{Xu_Peng_Cheng_Willis_Ritchie_2021} infer CAD modeling sequences from parametric solid shapes. Still, these methods are far from our objective which is to generate 3D CAD files based on image input.

\noindent{\textbf{Sequantial CAD Generation. }}
    Sequential CAD Generation involves utilizing sequences of modeling operations stored in parametric CAD files as supervision to train generative models \cite{ExtrudeNet, Wu_Chang_Zheng_2021, HNC_CAD, Xu_Willis_Lambourne_Cheng_Jayaraman_Furukawa_2022, Zhou_Tang_Zhou_2023}. The closest conditional generation work to our work is \cite{Li_Pan_Bousseau_Mitra_2020} and \cite{li2022free2cad}, which treat the conditional generation as the sequence-to-sequence generation, which trained a neural network on synthetic data to translate 2D user sketches (in stroke sequence) into CAD operations. However, these methods still need a substantial amount of user input and still cannot deal with the RGB image input as widely used existing image-to-3D works. This work mitigates the research gap by providing the first image-to-3D framework represented by a CAD command sequence.  

\begin{figure*}[htb]
\centering
\includegraphics[width=0.8\textwidth]{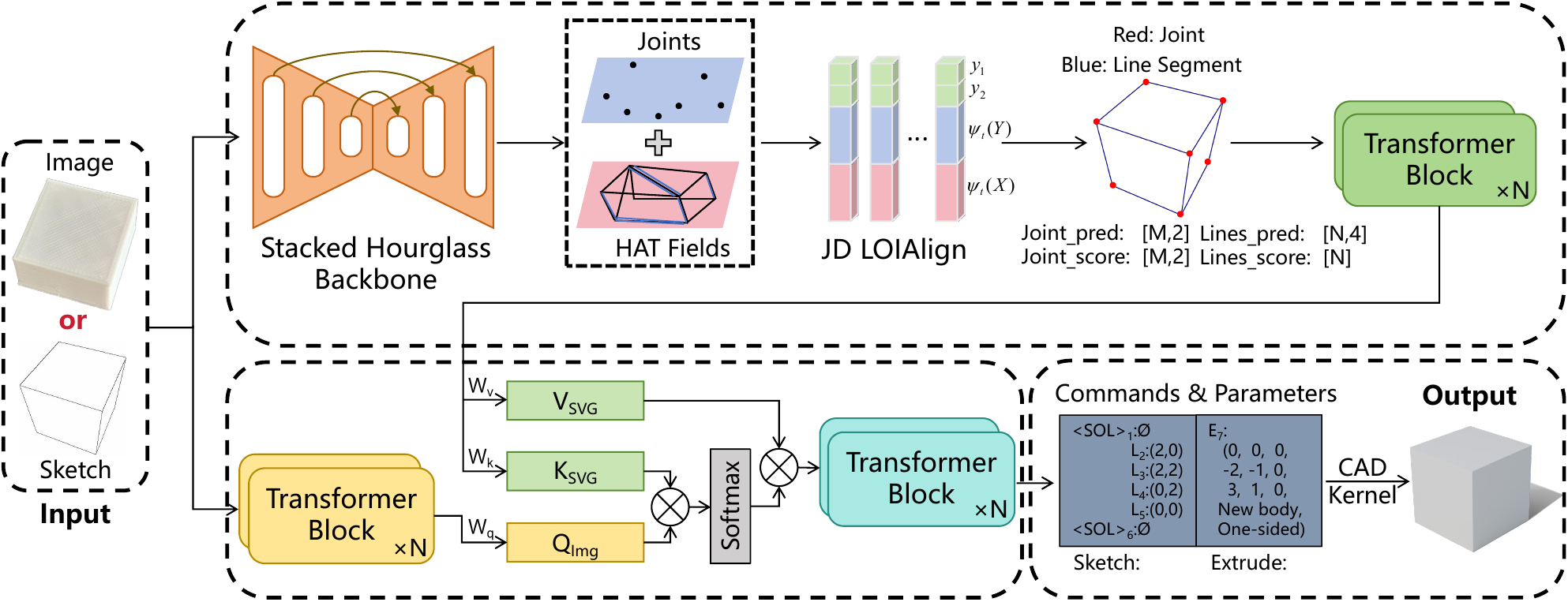}
\caption{Our method takes the input of images or sketches and uses a feature extractor with conditions added via extracted wireframe information to generate the command and parameters of a 3D CAD model. A CAD kernel can be used to convert the commands and parameters to a 3D model.} \label{Figzhu}
\vspace{-0.3cm}
\end{figure*}

\section{Preliminary: Specification of CAD Commands}
    The complete CAD toolkit supports a rich set of commands, although only a small fraction of them are commonly used in practice. Here, following previous works \cite{Wu_Chang_Zheng_2021}, we consider a subset of frequently used commands listed in Table \ref{tab2}. 
\begin{table}[h]
\centering
\caption{CAD commands and parameters used in Img2CAD}\label{tab2}
\setlength{\tabcolsep}{4mm}
\resizebox{0.4\textwidth}{!}{
\begin{tabular}{cc}
\hline
Commands    &       Parameters \\
\hline
$<SOL>$     &       $\phi$  \\
\hline
L           & \multirow{2}{*}{$x$, $y$ : line end-point} \\
(Line)      &   \\
\hline
\multirow{2}{*}{A}          & $x$, $y$ : arc end-point \\
\multirow{2}{*}{(Arc)}      & $\alpha$ : sweep angle \\
                            & $f$ :counter-clockwise flag \\
\hline
R                           &       $x$, $y$ : center \\
(Circle)                    &       $r$ : radius    \\
\hline
\multirow{6}{*}{E}          &       $\theta$, $\phi$, $\gamma$ : sketch plane orientation \\
\multirow{6}{*}{(Extrude)}  &       $p_{x}$,$p_{y}$,$p_{z}$ : sketch plane origin \\
                            &       $s$ : scale of associated sketch profile \\
                            &       $e_{1}$, $e_{2}$ : extrude distances toward both sides \\
                            &       $b$ : boolean type \\
                            &       $u$ : extrude type \\
\hline
$<EOS>$                     &       $\phi$  \\
\hline
\end{tabular}}
\end{table}
    These commands are categorized into two types: sketch and extrusion, and processes sufficient expressive power.
    \noindent\textbf{Sketch:} In CAD software, each closed curve is referred to as a loop, and one or more loops form a closed region called a profile. Therefore, a profile is described by a series of loops on its boundary, each loop begins with the indicator command $\langle SOL \rangle$ followed by a series of curve commands $C_{i}$. In practice, we consider the three most commonly used curve commands: draw line, arc, and circle. Each curve command $C_{i}$ is described by its curve type $t_{i} \in \{\langle SOL \rangle, L, A, R\}$ and parameters as listed in Table \ref{tab2}. The curve parameters specify the 2D position of the curve in the sketch plane for a local reference frame, while its position and orientation in 3D will be described in the relevant extrusion command. In summary, a sketch profile $S$ is composed of a series of loops, where each loop $Q_i $ contains a series of curve commands starting from the command $\langle SOL \rangle$, and each curve command $C_{j} = (t_j, p_j)$ specifies the curve type $t_i$ and shape parameters $p_{j}$. \noindent\textbf{Extrusion:}The extrusion command in CAD modeling has two main functions: converting a 2D sketch into a 3D entity with options like single-sided, symmetric, or double-sided extrusions, and defining how the extruded shape interacts with existing shapes, allowing for operations like union, subtraction, or intersection via parameter $b$. This command also necessitates defining the three-dimensional orientation of the sketch plane and its two-dimensional local reference frame. This is achieved through a rotation matrix defined by parameters $(\theta, \gamma, \phi)$ in Table \ref{tab2}, aligning the world reference frame with the plane's local reference frame and aligning the z-axis with the plane's normal direction. In essence, CAD models are described as a sequence of curves and extrusion, where each command $\mathit C_i$ consists of a command type $\mathit t_i$ and corresponding parameters $\mathit p_i$.

\begin{figure*}[h]
\centering
\includegraphics[width=0.6\textwidth]{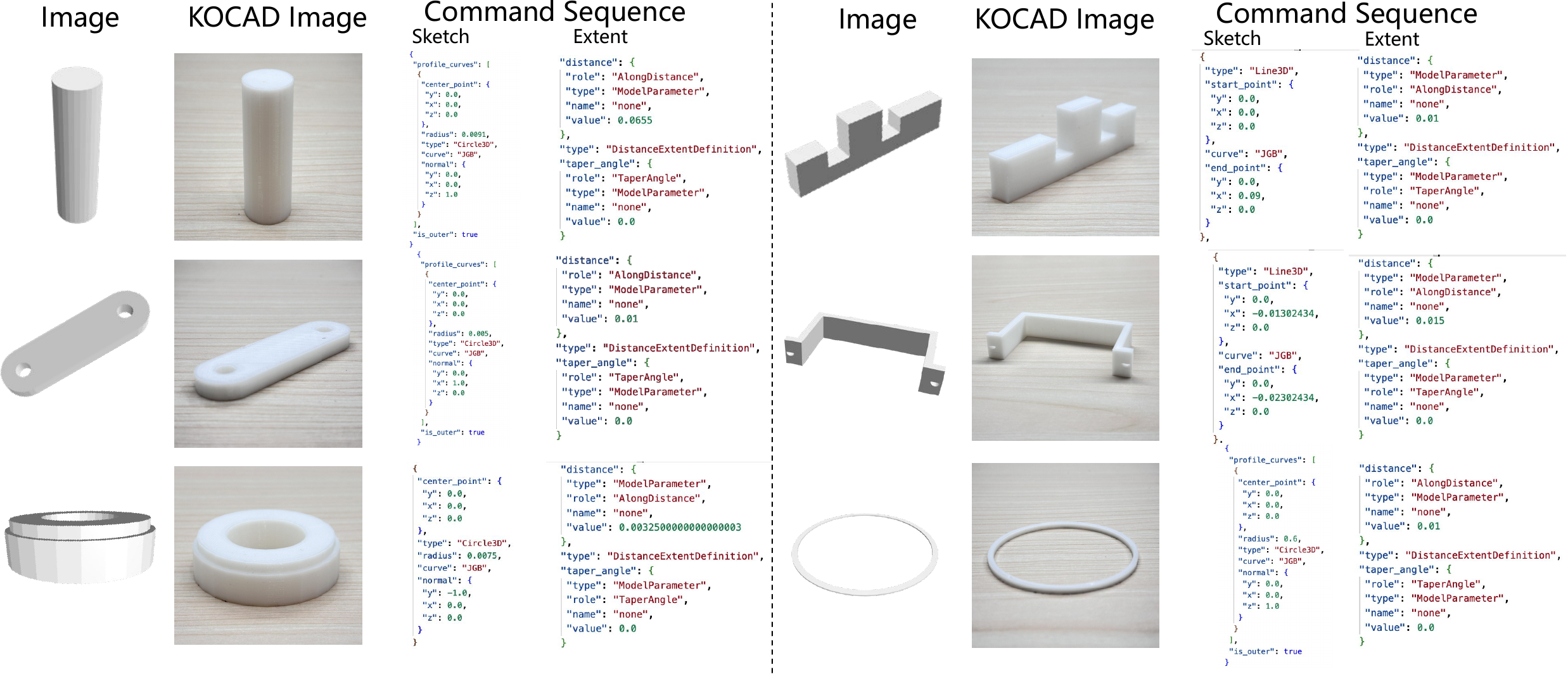}
\caption{The visualization of the KOCAD dataset and the corresponding command sequences.} \label{Fig2}
\vspace{-0.5cm}
\end{figure*}

\section{Method}
\label{ssec:4.0}
    Our proposed Img2CAD is shown in Fig. \ref{Figzhu}. The architecture remains consistent whether handling sketch or image input, ensuring flexibility without requiring modifications to its structure. First, we employ a robust feature extractor to the input. In this case, we opt to utilize a pre-trained feature extractor \cite{oquab2023dinov2} based on the Vision Transformer (ViT) architecture. 

    In our experiment, we find that only using a ViT encoder still cannot extract enough information due to the huge domain gap between our input and the output sequence. Therefore, we need to leverage additional features to accurately obtain the object's geometric information. Drawing inspiration from primate visual systems, researchers have long utilized geometric information such as salient points, line segments, and planes to depict image contents effectively, particularly in tasks requiring geometric understanding. Among these geometric representations, we have chosen to use vectorized wireframes, which capture line segments and their associated endpoints (primarily junctions), for their representations of the underlying boundary structures of objects and generic regions \cite{xue2023holistically}. This is also close to the form of a parametric CAD sequence with a strong emphasis on vectorized boundaries. 
   
    Inspired by previous work in wireframe extraction \cite{xue2023holistically}, we first utilize a Stacked Hourglass Network \cite{Newell_Yang_Deng_2016} as the backbone to extract feature maps that correspond to the line data represented as a line field $\hat{L}_{n}$ and Endpoint Proposals $\hat{P}_{m}$. The predicted line segment proposals and endpoints are bonded and form an initialized wireframe.    
    Specifically, in $\hat{L}_{n}$, we find the nearest endpoint proposals $\hat{P}_{m}$ for the two endpoint proposals of the line segment proposal $\hat{L}_{n}$, denoted as $x_{1}$ and $x_{2}$. Similarly, find the nearest endpoint proposal $x_{1}$ for the endpoint proposal $y_{1}$, and likewise, find the nearest endpoint proposal $x_{2}$ for the endpoint proposal $y_{2}$. We calculate the squared Euclidean distances between them as $\delta_{1}$ and $\delta_{2}$. The maximum distance is defined as $\delta=Max(\delta_{1}, \delta_{2})$. Smaller distances indicate higher quality for the line segment $\hat{L}_{n}$ (a higher likelihood of collinearity). Simultaneously, use a threshold $\varepsilon$ to select high-quality line segment proposals whose binding cost $\delta$ is less than this threshold. Finally, generate a new set of endpoint-enhanced line segment proposals $\hat{L}_{n}=(x_{1},x_{2},y_{1},y_{2})$, consisting of line segments $\hat{L}$.

    Since points $x_{1}$ and $x_{2}$ on the line segments are considered "background" points, they are not validated even after binding. Hence, Line-of-Interest (LOI) Pooling is introduced to validate line segment proposals by connecting the entire line segment proposal with data evidence. The sampling function $\Psi_{t}$ is used to map background points to a point on the line segment in the following manner: $
        \mathit \Psi_{t}(X) = (1-t) \cdot x_{1} + t \cdot x_{2}
    $, where $X$ is the mapped point, $\mathit t \in [0,1]$.

    For LOI, the model maintains three sets of sampling points: (1) the two endpoints {$y_{1}$, $y_{2}$}; (2) the center point $\Psi_{t}(X)$ of $x_{1}$ and $x_{1}$; (3) the center point $\Psi_{t}(Y)$ of $y_{1}$ and $y_{1}$. By decoupling the endpoints and middle points, the model gains geometric awareness when learning to validate proposals, capturing the co-occurrence relationship between line segment proposals and node-guided deduced line segment proposals.

    After the above operation, the coordinates of points are denoted as $L_{n} \in \mathbb{R}^{N \times 2}$. To efficiently integrate with the line segment features $F^{C}_{j}$, we designed a point encoder to encode the coordinate points. The detailed design is as follows:

    Firstly, we use Embedding to encode them into 512-dimensional features. Then, we use the encoder part of the transformer, designed as encoder layers, to further extract features from the encoded features. Eventually, we condition the obtained line and joint feature to the feature extracted by the original image encoder via the cross-attention mechanism. The condition enables the wireframe features to guide the fusion of sketch features. Since sketch features $F^{S}_{j}$ struggle to effectively distinguish between foreground and background features during extraction, we use them as queries to retrieve key point information related to sketch features from point features $F^{P}_{j}$. This approach increases the distance between foreground and background features, thereby facilitating the decoder's subsequent work.

    Similar to the point encoder, our decoder is also constructed based on Transformer blocks, using the same hyperparameter settings as the encoder. It takes the feature $F_{j}$ as input and feeds the output of the last Transformer block into a linear layer to predict the CAD command sequence $\hat{M}=[C_{1}, C_{2},..., C_{N_{C}}]$, including the command type $\hat{t}^{j}_{n}$ and parameters $\hat{z}^{j}_{n}$ for each command $C_{i}$. The formula is as follows:
\begin{align}
    \begin{split}
        \hat{p}_{j}(\hat{t}^{j}_{i},\hat{z}^{j}_{i}) &= Decoder(F_{j})
    \end{split}
\end{align}
where $F_{j}$ represents the fused features of the j viewpoint under the i CAD model; $\hat{t}^{j}_{i} \in \mathbb{R}^{N \times 4}$ represents that there are j commands in the i- CAD model, each command having four command forms: Line, Arc, Circle, Extrude; $\hat{z}^{j}_{i} \in \mathbb{R}^{N \times 4 \times 16}$, each command form has 16 command parameters, $z=[x,y,\alpha, f, r, \theta, \gamma, p_{x}, p_{y}, p_{s}, s, e_{1}, e_{2}, b, \mu]$.

\section{Experiments}
\label{ssec:5.0}

\subsection{The Synthetic Data and the ABC-mono Dataset}
\label{ssec:5.1}
   There was a lack of existing datasets that contained paired images and CAD models represented in both sketch and extrude formats since most existing 3D generation datasets only provide GT 3D models in mesh representation. Here, we build a customized dataset, ABC-mono. The CAD models used in our research were sourced from the DeepCAD dataset \cite{Wu_Chang_Zheng_2021} as well as from our self-collected data. The total number of 3D models is 208,853. They were divided into 90\% for training, 5\% for validation, and 5\% for testing. Following Willis et al. \cite{willis2021engineering}, duplicate sketch and extrude subsequences and any invalid sketch-and-extrude operations were removed. After obtaining the CAD models, we render the model in Blender using random viewpoints. For each CAD model, we render 36 images with the viewpoint information also saved, which results in 7,518,708 samples. 

    Then, we extract the synthetic sketch information from the rendered images. We first apply Gaussian smoothing filters to the images to reduce noise and improve image quality. Then, the gradients and gradient directions of the images are calculated, followed by Non-Maximum Suppression (NMS) to eliminate non-edge pixels, retaining only some thin lines as the candidate strokes. Finally, the edge detection process is completed by applying high and low thresholding and connecting edges. 
\vspace{-0.1cm}
\subsection{The Real Data and the KOCAD Dataset}
\label{ssec:5.2}
    It is worth noting that the rendered images lack realistic materials and still have a huge gap in real-life applications with varied surface properties and lighting conditions. Therefore, we create a dataset of real object images and the corresponding sketch and extrude sequence of the CAD model. We select around 100 objects in the ABC-mono dataset and have the objects accurately fabricated using commercial 3D printers (KOKONI SOTA Combo 3D printer, calibrated dimensional accuracy \textless 0.2mm) by multiple materials with slightly different surface properties (PLA plastic, ABC plastic, and PC plastic). We capture the image of the objects in varying lighting conditions and backgrounds, which results in the KOCAD dataset, which comprises 300 images. 
    
\begin{figure}[h]
\centering
\includegraphics[width=0.3\textwidth]{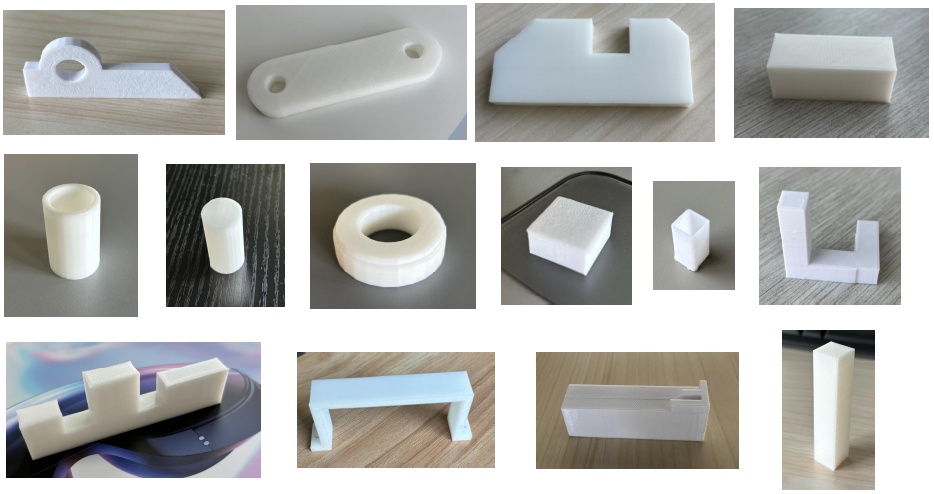}
\caption{More Examples of KOCAD Dataset. Objects are fabricated with different materials and captured under varying environmental conditions.} \label{Fig2}
\end{figure}
\vspace{-0.2cm}

\subsection{Implementation Details and Evaluation Metrics}
\label{ssec:5.3}
    The input size of the model framework for image inputs was set to 512 $\times$ 512 $\times$ 3, with a batch size of 1024. The training was performed using the Adam optimizer on 4 NVIDIA A100 GPUs, with a learning rate of 0.001. We applied a dropout rate of 0.1 to all Transformer blocks and employed gradient clipping with a value of 1.0 during backpropagation.
    For training, we use the cross-entropy loss between the predicted CAD model and the ground truth model as:
\begin{align}
    \begin{split}
        \mathcal L = \sum^{M}_{i=0} (\sum^{N_{C}}_{j=0}\tau(\hat{t}^{j}_{i}, {t}^{j}_{i}) + \lambda\sum^{N_{C}}_{j=0}\tau(\hat{z}^{j}_{i}, {z}^{j}_{i}))
    \end{split}
\end{align}
    where $\tau(\cdot, \cdot)$ represents the standard cross-entropy, $\lambda$ represents the balance weight between the two terms ($\lambda = 2$).

    Following previous work \cite{Wu_Chang_Zheng_2021}, we utilized Command accuracy ($Cmd_{ACC}$), Parameter accuracy ($Param_{ACC}$), and the Invalid Ratio. Additionally, we employed the Chamfer Distance ($CD$) for shape fidelity measurement. 
    
\vspace{-0.2cm}
\subsection{Results in ABC-mono dataset}
\label{ssec:5.5}
\begin{figure}[h]
\centering
\includegraphics[width=0.5\textwidth]{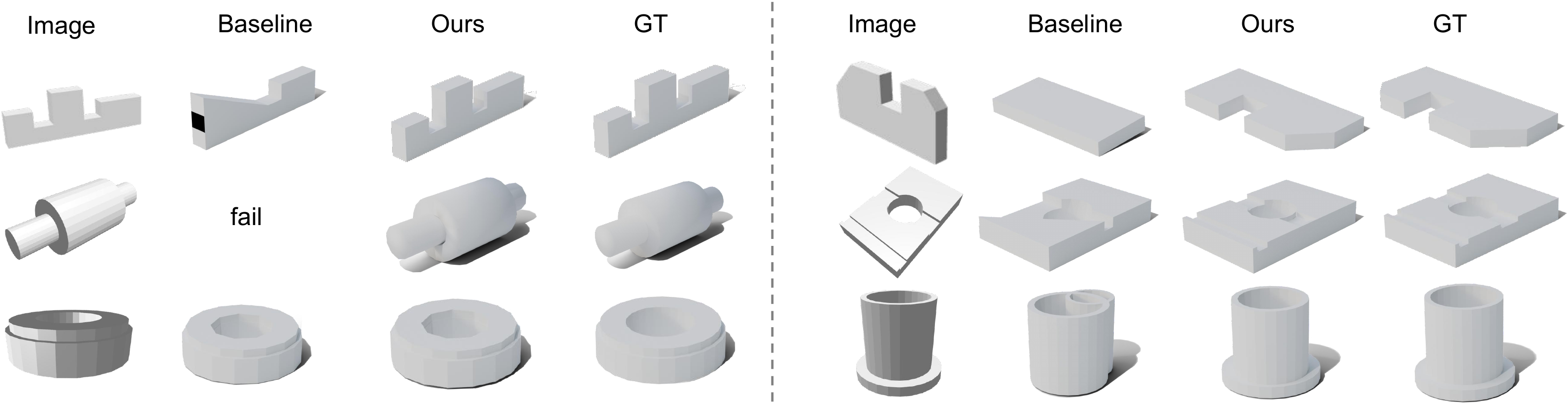}
\caption{Visualization of results with image input. } 
\vspace{-2mm}
\label{Imgin}
\end{figure}

    \noindent\textbf{Generation with Image Input.} Since there are no existing approaches that can handle sketch and extrude sequence generation from images, we build a baseline modified from DeepCAD \cite{Wu_Chang_Zheng_2021}. We replace the original command input with the image input and keep other settings unchanged (Denoted as "DeepCAD*"). We also build a modified "HNC-CAD*" in the similar manner \cite{HNC_CAD}. Experimental shows that our method, with the involvement of structured visual geometry learning, is capable of producing high-quality and high-fidelity 3D shapes, with higher command accuracy and parameter accuracy for synthetic data as in Tab. \ref{tab4}. The effectiveness can also be validated by qualitative comparison as in Fig. \ref{Imgin}. 

\vspace{-0.3cm}
\begin{table}[h]
\centering
\caption{Quantitative Comparison for Image Input.}\label{tab4}
\setlength{\tabcolsep}{1.0mm}
\begin{tabular}{ccccc}
\hline
   & $Cmd_{ACC}$↑ & $Param_{ACC}$↑ & Invalid ratio↓ & $CD$↓ \\
\hline
DeepCAD* & 0.77640 & 0.66027 & 0.31959 & 0.22564\\
\hline
HNC-CAD* & /  &  / &  0.68084 & 0.61855\\
\hline
\textbf{Ours} & \textbf{0.80574} & \textbf{0.68773} & \textbf{0.28815} & \textbf{0.16144} \\
\hline
\hline
TripoSR\cite{TripoSR2024} & /  &  / &  / & 0.72065\\
\hline  
One-2-3-45\cite{liu2023one2345} & /  &  / &  / & 0.53707\\
\hline  
\end{tabular}
\end{table}
\vspace{-0.3cm}

\begin{figure}[h]
\centering
\includegraphics[width=0.51\textwidth]{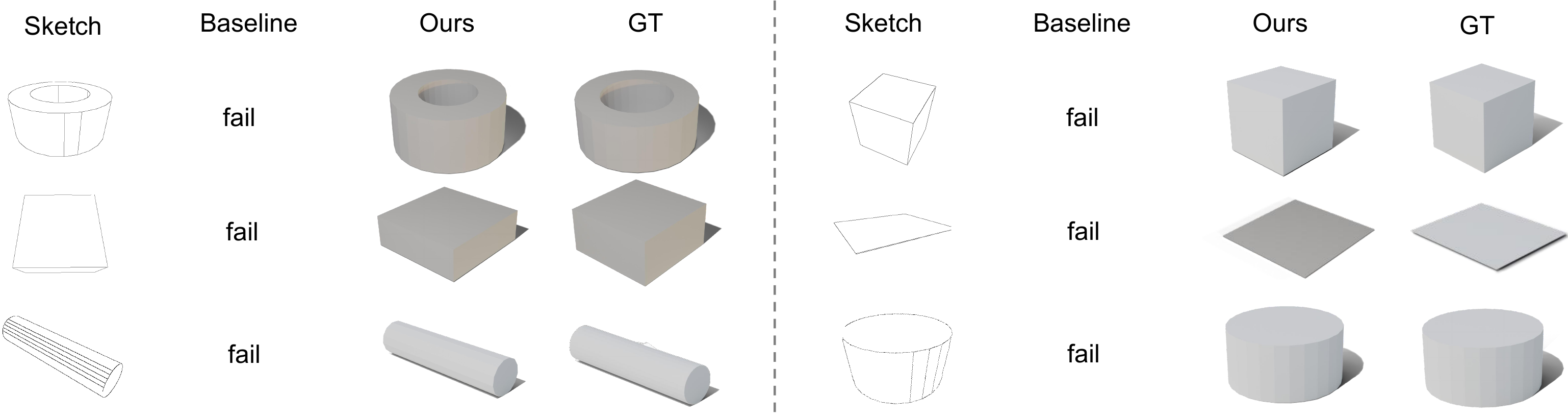}
\caption{Visualization of generation results with sketch input. The baseline method fails for 99.97\% of testing samples.} 
\label{sketchin}
\vspace{-0.3cm}
\end{figure}

\noindent\textbf{Generation with Sketch/Edge Map Input.}
    As sketching is one of the most natural ways we human design, we choose to test the sketch input. This is challenging because, unlike images, sketch input has no depth or texture information. The inherent sparsity and ambiguity of sketches pose additional challenges. The challenges are demonstrated in our experiment. When using the sketch to replace the image input, the invalid ratio becomes higher, and the command accuracy and parameter accuracy drop. It is worth noting that the baseline method (the modified DeepCAD*) fails to produce any meaningful result with the invalid ratio of 99.975\% shown in Table.\ref{tab3}, while our method, with the involvement of structured visual geometric learning, is still capable of producing meaningful results as shown in Fig. \ref{sketchin}. 

\begin{table}[htbp]
\centering
\caption{Quantitative Comparison for 3D Generation with Sketch/Edge Input.}\label{tab3}
\setlength{\tabcolsep}{1.8mm}
\begin{tabular}{ccccc}
\hline
       & $Cmd_{ACC}$↑ & $Param_{ACC}$↑  & Invalid ratio↓ & $CD$↓ \\
\hline
Baseline & 0.51286 & 0.42116 & 0.99975 & 0.97490 \\
\hline
\textbf{Ours} & \textbf{0.72844} & \textbf{0.60456} & \textbf{0.50202} & \textbf{0.37906} \\
\hline
\end{tabular}
\end{table}
\vspace{-0.5cm}

    \noindent\textbf{{Validated Multi-View Consistency.}}  A major challenge in 3D generation is ensuring multi-view consistency, especially when supervision comes from abstract commands and parameters rather than 3D structures. Without direct 3D supervision, models risk generating inconsistent outputs across different views, leading to geometric errors. To prevent this, we designed our training to promote consistency across multiple perspectives. We render 36 views of each object, encoding them into a global feature code used to generate CAD commands. The model learns to maintain a unified representation, ensuring consistent, accurate outputs across viewpoints. This strategy has proven effective, as validated empirically. We constructed a test set comprising 36 views for each 3D object. We evaluated the model's performance on images taken from different angles. The results, as depicted in the figure below, demonstrate that these metrics remain consistent across various viewpoints. Additionally, we conducted an ANOVA test, which showed no significant correlation between the viewpoint and the qualitative metrics (p\textless0.01). This shows that our method effectively preserves multi-view consistency.

\begin{figure}[h]
  \centering
   \includegraphics[width=0.8\linewidth]{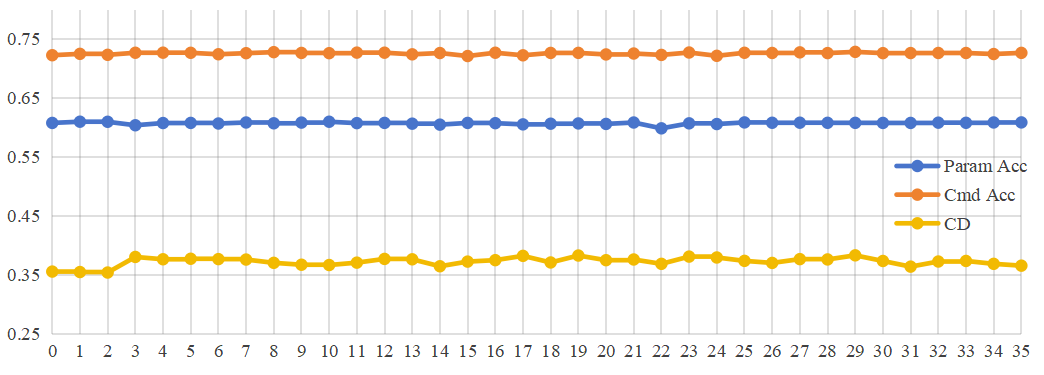}
   \vspace{-1mm}
  \caption{Result for varied viewpoints. X-axis: Viewpoint ID.}
  \vspace{-2mm}
\end{figure}

\begin{figure*}[h]
\centering
\includegraphics[width=\textwidth]{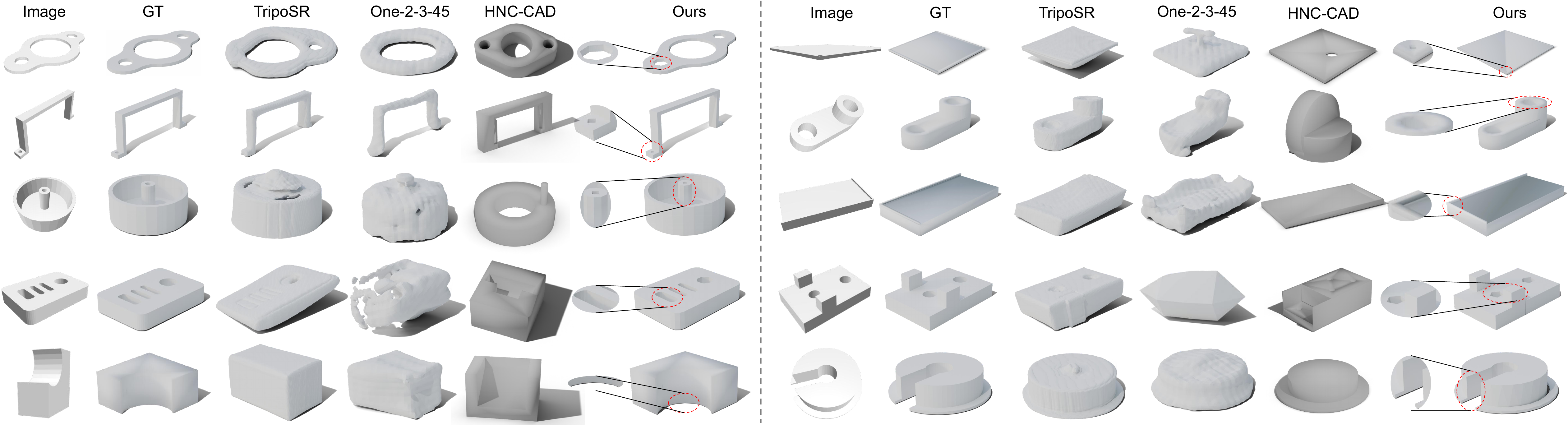}
\caption{Visualization for our method compared with recently popular 3D AIGC approaches. It is evident that our method is capable of producing 3D shapes with higher fidelity with more accurate details, cleaner surfaces, and sharper edges.} 
\label{Fig6}
\vspace{-0.5cm}
\end{figure*}

\noindent\textbf{Advantage of Using "Sketch and Extrude Representation" for 3D Generation.}
    As mentioned in the Introduction, the generation using the sketch and extruded representation can bring high-fidelity, high-quality, and compact 3D models, especially for man-made objects. We use the ABC-mono dataset and feed the image input into some popular and recently proposed image-conditioned 3D generation approaches. The Table. \ref{tab4}, it can be seen that our method is capable of producing higher-fidelity 3D models with lower CD. As shown in Fig. \ref{Fig6}, our method can produce compact, smooth/flat surfaces and clear edges, while existing approaches demonstrate lower visual qualities and more artifacts.

\noindent\textbf{{Speed Comparison.}}
    In table. \ref{speed}, we compare the computational efficiency of our proposed method with three state-of-the-art 3D model generation techniques. This substantial speed advantage is particularly beneficial for human-computer interaction and significant cost savings.

\begin{table}[h]
\centering
\caption{Inference Time comparison.}
\setlength{\tabcolsep}{1.0mm}
\resizebox{0.5\textwidth}{!}{  
\begin{tabular}{cccccc}
\hline
   & Wonder3D\cite{Wonder3D} & TripoSR\cite{TripoSR2024} & One-2-3-45\cite{liu2023one2345} & HNC-CAD*\cite{HNC_CAD} & {Ours} \\
\hline
Time↓ & 471.47s & 2.62s & 155.58s & 3.94s & {0.66s} \\
\hline  
\end{tabular}
}
\label{speed}
\vspace{-0.3cm}
\end{table}

\vspace{-0.3cm}

\subsection{Performance in Real-Life Scenarios}
\label{ssec:5.6}
    
\begin{figure}[h]
\centering
\includegraphics[width=0.3\textwidth]{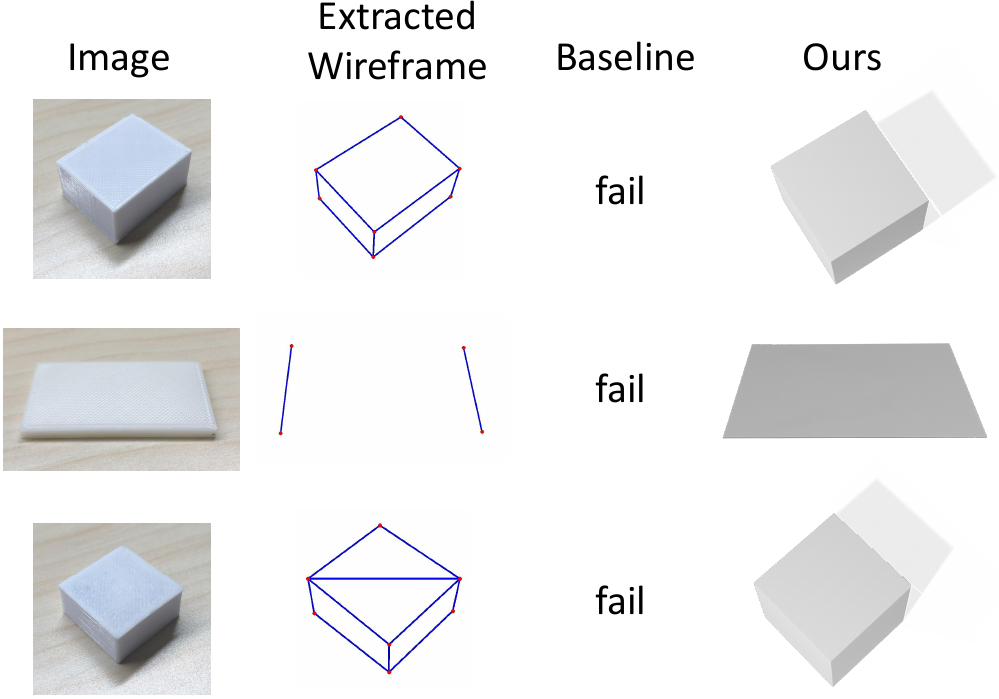}
\caption{Visualization on 3D Generation with Real-Life Image Input in KOCAD dataset. The baseline method reports 100\% failure while our proposed method is still capable of producing viable 3D shapes with the extracted wireframe.} 
\label{fig8}
\vspace{-0.3cm}
\end{figure}

    To verify the generalization capability of our method in practical applications and to better meet real-world needs, we tested our approach on our proposed real-life dataset, KOCAD. The varying lighting conditions and surface properties in this dataset pose significant challenges for 3D reconstruction. This challenge was evident in our baseline experiment, where the baseline method completely failed (100\% invalid). In contrast, our method produced viable results in over 65\% of cases. Visualization results further confirm that our method, supported by the extracted wireframe, is capable of generating plausible 3D models (Tab. \ref{tab5} and Fig. \ref{fig8}).
    
    The industrial value of this lies in the fact that, although the generated CAD models may not be fully precise, they serve as an excellent starting point for rapid prototyping. It allows designers or engineers to rapidly iterate and refine the model and save costs.

\begin{table}[h]
\centering
\caption{Performance Comparison on KOCAD.}\label{tab5}
\setlength{\tabcolsep}{2mm}
\begin{tabular}{ccccc}
\hline
    & $Cmd_{ACC}$↑ & $Param_{ACC}$↑ & Invalid ratio↓ & $CD$↓ \\
\hline
Baseline & 0.41128 & 0.44760 & 1.0 & / \\
\hline
\textbf{Ours} & \textbf{0.51302} & \textbf{0.48558} & \textbf{0.34711} & \textbf{1.49657} \\
\hline
\end{tabular}
\vspace{-0.5cm}
\end{table}

\vspace{-0.3cm}
\subsection{Ablation Study for the Usefulness of SVG}
    In our ablation study, we aimed to isolate the core contribution of our proposed method—Structured Visual Geometry (SVG) learning—and demonstrate its significant impact on improving image-to-CAD generation results. To evaluate this, we conducted two key experiments. First, we completely removed the SVG module from the pipeline, and second, we randomly discarded 50\% of the line and joint information from the SVG during training (Partial SVG Guidance). We performed these ablation experiments on both sketch/edge input and image input, and the trend remained consistent across both types of inputs. The results, summarized in the table \ref{toop}, highlights that the use of explicit geometric representations significantly enhances the model’s ability to decode accurate CAD commands and parameters.

\begin{table}[h]
\centering
\caption{Ablation Study Confirmed the Effectiveness of SVG.}
\scalebox{0.9}{  
\setlength{\tabcolsep}{1.0mm}
\begin{tabular}{ccccc}
\hline
 Image Input  & $Cmd_{ACC}$↑ & $Param_{ACC}$↑ & Invalid ratio↓ & $CD$↓ \\
\hline
Ours w/o SVG Guidance & 0.77640 & 0.66027 & 0.31959 & 0.22564\\
\hline
Ours w/ partial SVG Guidance & 0.78014 & 0.67885 & 0.30728 & 0.20277\\
\hline
\textbf{Ours} & \textbf{0.80574} & \textbf{0.68773} & \textbf{0.28815} & \textbf{0.16144} \\
\hline
\hline
Sketch Input  \\
\hline
Ours w/o SVG Guidance & 0.51286 & 0.42116 & 0.99975 & 0.97490 \\
\hline
Ours w/ partial SVG Guidance & 0.71550 & 0.59731 & 0.42641 & 0.50882 \\
\hline
\textbf{Ours} & \textbf{0.72844} & \textbf{0.60456} & \textbf{0.50202} & \textbf{0.37906} \\
\hline
\end{tabular}
}\label{toop}
\vspace{-0.5cm}
\end{table}

\subsection{Applications: High-Quality Realistic Rendering}
\begin{figure}[h]
    \centering
    \includegraphics[width=0.95\linewidth]{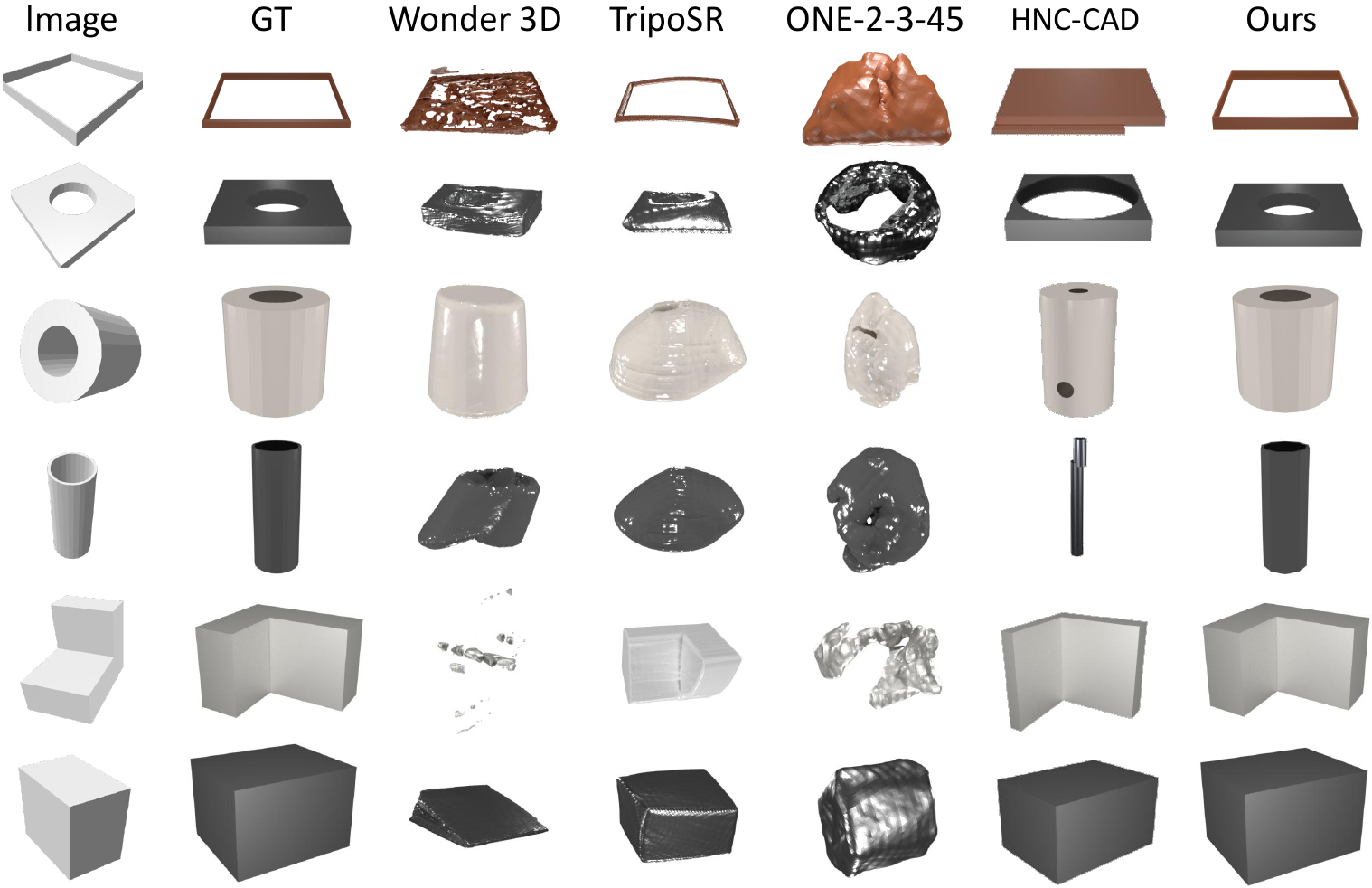}
    \caption{Superior rendering quality can be achieved with higher quality models from our method.}
    \label{sup_fig3}
\end{figure}

    We show an application by manually putting materials to the model and performing realistic rendering. Our findings reveal significant enhancements in the visual quality and realism of the generated 3D models by our method with CAD representation. As shown in Fig. \ref{sup_fig3}. 

\vspace{-0.5cm}
\section{Discussions and Limitations}
\label{ssec:6.0}
    This work introduces a novel approach to 3D generation using CAD commands. While the generated models can be integrated into existing software, they are rough prototypes requiring further refinement by human experts for precision tasks like manufacturing. These rough outputs are valuable for rapid prototyping, as they save time in design processes.

    Moreover, the CAD commands used can be seen as a language, opening opportunities to leverage large language models (LLMs) in future research. Early attempts have explored this, and integrating LLMs could enhance generation. Additionally, generating synthetic CAD models and corresponding sketches could create larger datasets to improve the model’s performance and accuracy.
    
    Despite its promise, the current method is limited to basic operations like sketching and extruding. Expanding the framework to handle more complex operations and building larger, more diverse datasets would enable the generation of intricate designs. Addressing these limitations will lead to more advanced 3D generation techniques in the future.
\vspace{-0.2cm}
\section{Conclusion}
\label{ssec:7.0}
    This paper addresses challenges in 3D AIGC, such as interpretability, editability, and surface quality, by introducing Img2CAD, a novel method for image-conditioned 3D generation using Sketch and Extrude Sequence representation. Img2CAD leverages Structured Visual Geometry (SVG) for CAD sequence reconstruction. Combining wireframe information with image features, significantly improves performance, especially with challenging sketch inputs and real-world images. Img2CAD achieves SOTA results in fidelity, quality, and speed. Additionally, new datasets ABC-mono and KOCAD are introduced. We believe that our method has significant implications for various industrial applications.

\bibliographystyle{IEEEtran}
\bibliography{ref}

\end{document}